\documentclass{article}
\usepackage{spconf,amsmath,epsfig}

\usepackage{cite}
\usepackage{url}
\usepackage{lipsum}
\usepackage{xurl}
\usepackage{siunitx}
\usepackage{pifont}
\usepackage{graphicx}
\usepackage{caption}
\usepackage{subcaption}
\usepackage{upgreek}
\usepackage{amsfonts}
\usepackage{makecell}
\usepackage{multirow}
\usepackage{booktabs}
\usepackage{tabularx}
\usepackage{cleveref}
\crefname{Fig}{Figure}{Figure}
\usepackage[usenames,dvipsnames]{xcolor} 
\usepackage{colortbl}
\usepackage{algorithm}
\usepackage{pgffor}
\usepackage{algorithmicx}
\usepackage{algpseudocode} %
\usepackage[export]{adjustbox}
\usepackage[short]{optidef}
\usepackage{soul}
\usepackage{ifthen}
\usepackage{tikz}
\usepackage{pgfplots}
\usepackage{pgfplotstable}
\usepackage{mathtools}
\usepgfplotslibrary{fillbetween}

\title{Annotation Cost Efficient Active Learning for Content Based Image Retrieval}
\name{Julia Henkel{\normalfont\textsuperscript{1}}, Genc Hoxha{\normalfont\textsuperscript{1}}, Gencer Sumbul{\normalfont\textsuperscript{1}}, Lars M\"{o}llenbrok{\normalfont\textsuperscript{1,2}} and Beg\"{u}m Demir{\normalfont\textsuperscript{1,2}}}
\address{\textsuperscript{1}Faculty of Electrical Engineering and Computer Science, Technische Universit\"at Berlin, Germany\\
\textsuperscript{2}BIFOLD - Berlin Institute for the Foundations of Learning and Data, Germany}
\begin{document}
%
\maketitle
\begin{abstract}
Deep metric learning (DML) based methods have been found very effective for content-based image retrieval (CBIR) in remote sensing (RS). For accurately learning the model parameters of deep neural networks, most of the DML methods require a high number of annotated training images, which can be costly to gather. To address this problem, in this paper we present an annotation cost efficient active learning (AL) method (denoted as ANNEAL). The proposed method aims to iteratively enrich the training set by annotating the most informative image pairs as similar or dissimilar, 
while accurately modelling a deep metric space. This is achieved by two consecutive steps. In the first step the pairwise image similarity is modelled based on the available training set. Then, in the second step the most uncertain and diverse (i.e., informative) image pairs are selected to be annotated. Unlike the existing AL methods for CBIR, at each AL iteration of ANNEAL a human expert is asked to annotate the most informative image pairs as similar/dissimilar. This significantly reduces the annotation cost compared to annotating images with land-use/land cover class labels. Experimental results show the effectiveness of our method. The code of ANNEAL is publicly available at https://git.tu-berlin.de/rsim/ANNEAL.
\end{abstract}

\section{Introduction}

The development of efficient and effective methods for content-based image retrieval (CBIR) from large-scale remote sensing (RS) image archives is one of the growing research interests in RS. CBIR aims at searching the most similar images to a given query image based on their semantic content. Deep metric learning (DML) methods have recently triggered a significant performance gain for CBIR problems in RS\cite{Sumbul:2021, surveyCBIR2021}. DML methods characterize high-level RS image semantics through deep neural networks (DNNs) by learning a feature space where similar images are located close to each other and dissimilar images remain far apart. Most of the DML methods require a high number of annotated training images to accurately characterize RS image content. However, the collection of annotated training images can be time-consuming and costly.
To reduce annotation costs, active learning (AL) methods have shown to be successful for CBIR problems in RS~\cite{Feracatu_CBIRAL2007, ALDemir2015}. These methods aim at optimizing the training set with a small number of annotated images. This is done by iteratively selecting only the most informative images to be annotated from an archive. The informativeness of the samples is usually defined based on uncertainty and diversity criteria. An AL method that combines the uncertainty and diversity criteria in two steps is presented in \cite {Feracatu_CBIRAL2007}. In the first step, the most uncertain samples are identified based on margin sampling \cite{schohn2000less}. In the second step, the most diverse samples among the uncertain ones are selected based on distances among image features. In \cite {ALDemir2015}, a triple criteria AL (TCAL) method is presented  that simultaneously combines the uncertainty and diversity criteria with density criterion to capture the underlying sample distribution of the archive images.
The above-mentioned methods rely on hand-crafted features and conventional classifiers such as support vector machines, and thus they are not suitable to be directly used within DNNs. In addition, they assume that RS images are annotated with land-use/land cover (LULC) class labels, which can be complex and costly to gather especially when dealing with a large number of LULC classes~\cite{hu2020onebitsupervision, onebitactivequery2022}.  

To address these issues, in this paper we present an annotation cost-efficient AL method (denoted as ANNEAL). Our method aims to define an optimized training set in an efficient manner, while accurately modeling a deep metric space. This is achieved in two steps. In the first step, pairwise image similarity is modeled based on the available training set. In the second step, the most informative image pairs are selected.
Then, a human expert labels them as similar or dissimilar by answering a simple yes/no question, which is inspired by one-bit supervision labelling strategy \cite{hu2020onebitsupervision}.
By relying on such binary annotations, ANNEAL is capable of reducing the annotation cost of a sample to one bit from $\log _{2}C$ bits, which is the labelling cost for $C$ classes in terms of bits of information.
\section{Proposed Cost-Efficient AL Method}

Let $\mathcal{X}$ be an RS image archive and $\mathcal{T} = \{\boldsymbol{X}_i^\mathcal{T}\}^M_{i=1}$ be an initial training set  composed of $M$ RS image pairs from $\mathcal{X}$. For $\boldsymbol{I}^{i,\mathcal{T}}_1$ and $\boldsymbol{I}^{i,\mathcal{T}}_2 \in \mathcal{X}$, each pair $\boldsymbol{X}_i^\mathcal{T} = (\boldsymbol{I}^{i,\mathcal{T}}_1,\boldsymbol{I}^{i,\mathcal{T}}_2)\in \mathcal{T}$ is associated with a similarity label $\boldsymbol{y}_i^{\mathcal{T}}$, where $\boldsymbol{y}_i^{\mathcal{T}} = 1$ if $\boldsymbol{I}^{i,{\mathcal{T}}}_1$ and $\boldsymbol{I}^{i,\mathcal{T}}_2$ are similar images and $\boldsymbol{y}_i^{\mathcal{T}} = 0$ otherwise. Let $\mathcal{U}=\{ \boldsymbol{X}_p^{\mathcal{U}}\}^N_{p=1} $ be a set of $N$ unlabeled pairs where $N \gg M$.

\begin{figure}
    \centering
    \includegraphics[scale =0.19]{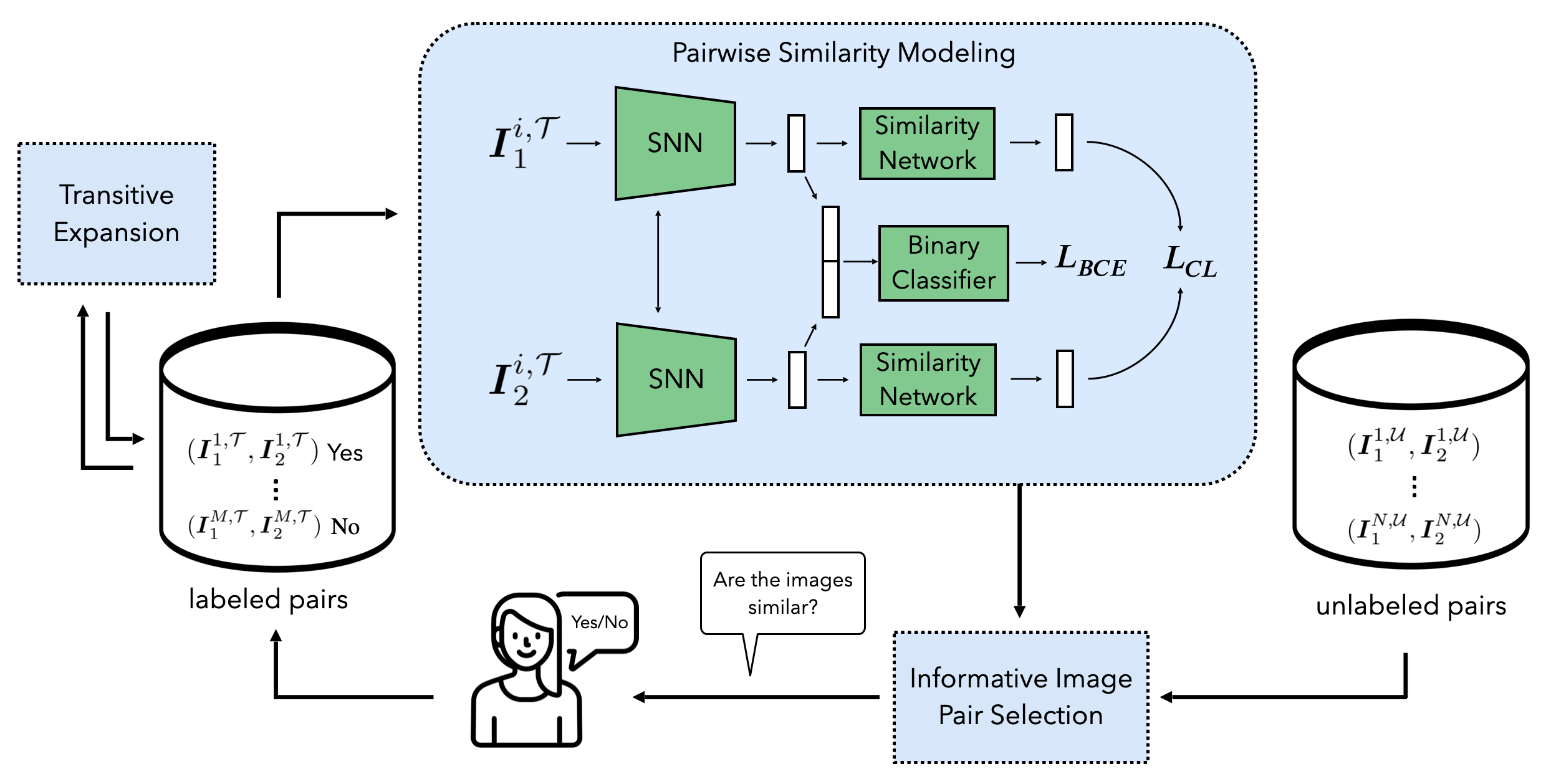}
    \caption{An illustration of the proposed method.}
    \label{Fig:ALscheme}
\end{figure}

The aim of the proposed method is to present a cost-efficient annotation process by selecting to label only the most informative image pairs in $\mathcal{U}$ as similar or dissimilar, while modelling a deep metric space. This is achieved in two steps, which are iterated in an AL fashion (see Fig.~\ref{Fig:ALscheme}). In \textit{(1) pairwise similarity modelling}, we use the current labeled training set $\mathcal{T}$ to model the similarity of the images present in $\mathcal{X}$. On one hand, we learn a metric space where similar images are located close to each other and dissimilar images far apart. On the other hand, we learn a binary classifier to classify images as similar or dissimilar. In \textit{(2) informative image pair selection}, we select the most informative image pairs from $\mathcal{U}$ to be labeled and added to the current labeled training set $\mathcal{T}$. The selection of the most informative pairs is based on uncertainty and diversity criteria. 
Finally, in \textit{transitive training set expansion}, we extend our currently labeled training set $\mathcal{T}$ with zero cost  exploiting that the similarity relation is naturally transitive \cite{roy2018exploiting}. We explain all steps in detail in the following subsections.

\subsection{Pairwise Similarity Modelling }
In this step, we aim to model the pairwise image similarity based on the current set $\mathcal{T}$ of labeled image pairs. 
To this end, we employ two DNNs with shared weights (i.e., Siamese neural network, SNN) that are followed by a binary classifier (BC) to classify images as similar or dissimilar. 
We feed each image pair $\boldsymbol{X}_i^{\mathcal{T}} \in \mathcal{T}$ into the Siamese network to obtain the corresponding image features~$(\boldsymbol{f}^{i,\mathcal{T}}_1,\boldsymbol{f}^{i,\mathcal{T}}_2)$. Then, the image features are concatenated and forwarded to the BC to obtain the similarity prediction $\boldsymbol{\hat{y}}_i^{\mathcal{T}}$ of the pair. 
To accurately model a metric space over the latent representations, we employ contrastive loss  \cite{contrastiveloss:2006} for learning its parameters as follows:
\begin{align}
    \!\!\mathcal{L}_{CL}\! &=\! \left\{\!\!\!
        \begin{array}{ll}
            1 - s(G(\boldsymbol{f}^{i,\mathcal{T}}_1), G(\boldsymbol{f}^{i,\mathcal{T}}_2)),\!\!\! & \boldsymbol{y}_i^{\mathcal{T}} = 1 \\
            \max{(0, s(G(\boldsymbol{f}^{i,\mathcal{T}}_1), G(\boldsymbol{f}^{i,\mathcal{T}}_2))\!-\!m}),\!\!\! & 
            \boldsymbol{y}_i^{\mathcal{T}} = 0
        \end{array}
    \right.
 \end{align}
where $m$ is the margin parameter, $s(\cdot, \cdot)$ represents the cosine similarity and $G(\cdot)$ is the similarity network composed of three fully connected layers. Next, the model parameters are further optimized in the BC by employing the binary cross entropy loss $\mathcal{L}_{BCE}$ between $\boldsymbol{y}_i^{\mathcal{T}}$ and $\boldsymbol{\hat{y}}_i^{\mathcal{T}}$ as follows:
\begin{align}
    \mathcal{L}_{BCE} &= \boldsymbol{y}_i^{\mathcal{T}}\log(\boldsymbol{\hat{y}}_i^{\mathcal{T}})+ (1-\boldsymbol{y}_i^{\mathcal{T}})\log(1-\boldsymbol{\hat{y}}_i^{\mathcal{T}}).
 \end{align} 
 The whole system is trained end-to-end with the overall loss $\mathcal{L}$, which is defined as follows:
 \begin{align}
    \mathcal{L} =  (1-\beta)\mathcal{L}_{CL}+\beta\mathcal{L}_{BCE},
 \end{align} 
 where $\beta$ is a balancing factor.  Once the network is trained on~$\mathcal{T}$, we perform CBIR by comparing the feature of the given query with that of each image in the archive $\mathcal{X}$.

\subsection{Informative Image Pair Selection }
 We select the $k$ most informative image pairs from $\mathcal{U}$ in two consecutive steps based on the criteria: 1) uncertainty and 2) diversity. In the first step, we select a set $\mathcal{S}_{\mathcal{U}}=\{\boldsymbol{X}_1^{\mathcal{U}},\ldots,\boldsymbol{X}_h^{\mathcal{U}}\}$ with $h>k$ of the most uncertain pairs for which the confidence of the BC in assigning the correct similarity label is low (i.e., similarity probability is close to $0.5$). 
 In the second step, we select the most diverse pairs among the most uncertain ones through clustering. Accordingly, we use k-means clustering algorithm to cluster the $h$ most uncertain pairs into~$k$ different clusters. Then, we select a set $\mathcal{S}_{\mathcal{U},\mathcal{D}}=\{\boldsymbol{X}_1^{\mathcal{U}},\ldots,\boldsymbol{X}_k^{\mathcal{U}}\}$ of the most uncertain pair per cluster, resulting in $k$ most informative image pairs. All pairs in $\mathcal{S}_{\mathcal{U},\mathcal{D}}$ are asked to be labelled by a human expert and added to the training set $\mathcal{T}$.  

\begin{figure}
\vspace*{-3mm}
     \centering
     \begin{subfigure}[c]{0.5\linewidth}
         \centering



\begin{tikzpicture}

\node[draw=gray!60, circle, fill = gray!10, scale = 0.7, minimum size = 40pt] (A) at (1.6,-0.16) {$I_1^{j,\mathcal{T}}$};
\node[draw=gray!60, circle, fill = gray!10, scale = 0.7, minimum size = 40pt] (B)  at (0,0.8) {$I_1^{i,\mathcal{T}}$};
\node[draw=gray!60, circle, fill = gray!10, scale = 0.7, minimum size = 40pt] (C)  at (1.6,1.6) {$I_2^{i,\mathcal{T}} \!= \!I_2^{j,\mathcal{T}}$};


\draw[dashed, color=blue!90!black, very thick] (A) -- (B);
\draw[-, color=blue!90!black, very thick] (B) -- (C);
\draw[-, color=blue!90!black, very thick] (A) -- (C);

\end{tikzpicture} 
        \caption{}
     \end{subfigure}%
     \begin{subfigure}[c]{0.5\linewidth}
         \centering
        \begin{tikzpicture}


\node[draw=gray!60, circle, fill = gray!10, scale = 0.7, minimum size = 40pt] (A) at (1.6,-0.16) {$I_1^{j,\mathcal{T}}$};
\node[draw=gray!60, circle, fill = gray!10, scale = 0.7, minimum size = 40pt] (B)  at (0,0.8) {$I_1^{i,\mathcal{T}}$};
\node[draw=gray!60, circle, fill = gray!10, scale = 0.7, minimum size = 40pt] (C)  at (1.6,1.6) {$I_2^{i,\mathcal{T}} = I_2^{j,\mathcal{T}}$};


\draw[dashed, color=orange!90!black, very thick] (A) -- (B);
\draw[-, color=blue!90!black, very thick] (B) -- (C);
\draw[-, color=orange!90!black, very thick] (A) -- (C);

\end{tikzpicture} 
        \caption{}
     \end{subfigure}
    \caption{An illustration of the zero-cost expansion of the training set by exploiting the transitivity property of similarity. 
    In~(a), $(\boldsymbol{I}_1^{i,\mathcal{T}}, \boldsymbol{I}_2^{i,\mathcal{T}})$ and $(\boldsymbol{I}^{j,\mathcal{T}}_1, \boldsymbol{I}^{j,\mathcal{T}}_2)$ are similar pairs and $\boldsymbol{I}^{i,\mathcal{T}}_2 = \boldsymbol{I}^{j,\mathcal{T}}_2$. Thus, $(\boldsymbol{I}^{i,\mathcal{T}}_1, \boldsymbol{I}^{j,\mathcal{T}}_1)$ is assumed to be a similar pair. In~(b), $(\boldsymbol{I}^{i,\mathcal{T}}_1, \boldsymbol{I}^{i,\mathcal{T}}_2)$ is a similar pair, $(\boldsymbol{I}^{j,\mathcal{T}}_1, \boldsymbol{I}^{j,\mathcal{T}}_2)$ is a dissimilar pair and $\boldsymbol{I}^{i,\mathcal{T}}_2 = \boldsymbol{I}^{j,\mathcal{T}}_2$. Thus, $(\boldsymbol{I}^{i,\mathcal{T}}_1, \boldsymbol{I}^{j,\mathcal{T}}_1)$ is  assumed to be a dissimilar pair.}
    \label{Fig:transitive}
\end{figure}
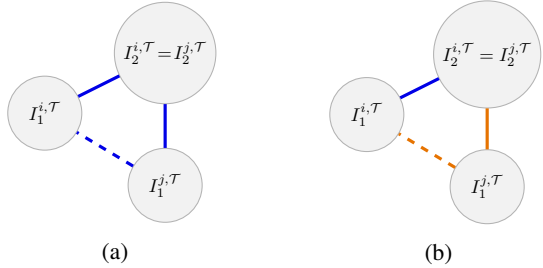

After adding the selected image pairs to $\mathcal{T}$, we further enrich our currently labeled training set with zero cost based on the transitive property of similarity \cite{roy2018exploiting}. The transitive operator is illustrated in Fig.~\ref{Fig:transitive} and defined as follows. Let $\boldsymbol{X}_i^{\mathcal{T}}$ and $\boldsymbol{X}_j^{\mathcal{T}} \in \mathcal{T}$ be two image pairs with $\boldsymbol{I}^{i,\mathcal{T}}_2 = \boldsymbol{I}^{j,\mathcal{T}}_2$. If $\boldsymbol{X}_i^{\mathcal{T}}$ and $\boldsymbol{X}_j^{\mathcal{T}}$ are similar pairs (i.e. $\boldsymbol{y}_i^{\mathcal{T}} = 1$ and $\boldsymbol{y}_j^{\mathcal{T}} = 1$), then we label pair $\boldsymbol{X}_k^{\mathcal{T}} = (\boldsymbol{I}^{i,\mathcal{T}}_1,\boldsymbol{I}^{j,\mathcal{T}}_1)$ as similar with $\boldsymbol{y}_k^{\mathcal{T}} = 1$. If $\boldsymbol{X}_i^{\mathcal{T}}$ is a similar pair and $\boldsymbol{X}_j^{\mathcal{T}}$ is a dissimilar pair (i.e. $\boldsymbol{y}_i^{\mathcal{T}} = 1$ and $\boldsymbol{y}_j^{\mathcal{T}} = 0$), then we label the pair $\boldsymbol{X}_k^{\mathcal{T}} = (\boldsymbol{I}^{i,\mathcal{T}}_1,\boldsymbol{I}^{j,\mathcal{T}}_1)$ 
as dissimilar with $\boldsymbol{y}_k^{\mathcal{T}} = 0$. The newly labeled pairs are added to the training set $\mathcal{T}$. Note that we only perform one transitive step as illustrated in Fig.~\ref{Fig:transitive} and do not proceed another transitive step with the newly generated pairs.


The three steps are repeated until we are satisfied with the  retrieval performance or a predefined labeling budged is reached.  
Following \cite{hu2020onebitsupervision}, we compute the cost of labeling in terms of bits of information. Annotating  $n$ samples (images) with $C$ classes  has an annotation cost of $n \log_2{C}$ bits of information. 
Since in ANNEAL we annotate pairs of images with similarity class label (i.e., similar/dissimilar), we reduce the annotation cost to  only $n \log_2{2} = n$ bits for $n$ pairs. This demonstrates the cost efficiency of the proposed method.  

\section{Experimental Results}
The experiments were conducted on UCM dataset that includes 2100 images, each of which is associated with one of 21 categories \cite{UCM_dataset}. We randomly split into training ($80\%$), validation ($10\%$), and test ($10\%$). 
We randomly select $5\%$ of the images from the training split of the UCM dataset and build the initial training  set $\mathcal{T}$ by randomly choosing four similar and four dissimilar images for each of the labeled images based on their category labels. The initial unlabeled training set $\mathcal{U}$ consists of all training pairs that are not included in $\mathcal{T}$.
At each AL iteration, $\mathcal{T}$ is enriched by adding $k = 336$ 
most informative pairs in $\mathcal{U}$. Thereby, $336$ bits of information are included in the newly labeled samples at each AL iteration. 
We choose ResNet18 as the backbone of the SNN and train with Adam optimizer and learning rate $10^{-4}$. The BC and the similarity network consist of three fully connected layers with hidden dimension $512$, $256$ and $256$. Based on cross-validation strategy, we set $k = 4 \cdot h$, $\beta = 0.1$ and $m = 0.1$.
Naturally, there are much more dissimilar than similar image pairs in UCM dataset.
Approximately $5\%$ of all pairs are similar and $95\%$ are dissimilar. Since this imbalance is also reflected in the training set $\mathcal{T}$, we stabilize the training by oversampling the minority class, as proposed in \cite{branco2016survey}.

We evaluate the CBIR performance by taking the query images from the validation set and performing the retrieval on the test set. 
We measure the performance in terms of mean Average Precision at five retrieved images (mAP@5) at every AL iteration.  To have statistically meaningful results, we repeat all experiments three times and average the scores. 
We study the effect of each individual AL criteria involved in ANNEAL and compare it with random pair selection (Random). Fig.~\ref{Fig:results2} shows the retrieval results in terms of MAP@5 and the cost in terms of bits of information. From the figure one can clearly see that ANNEAL yields the highest results outperforming random sampling with a significant margin (up to $30\%$) for any number of bits of information. In addition, we can see that by combining both criteria we are able to improve over ANNEAL-U and reach the accuracy of the fully training set at a lower cost ($2500$ bits of information). This demonstrates that the combination of the two criteria is crucial for efficiently building an optimized training set.

To evaluate the cost effectiveness of the proposed method, we compare it with the AL method for CBIR that relies on LULC classes \cite{ALDemir2015} denoted as TCAL. In addition to uncertainty and diversity criteria, TCAL incorporates the density criteria. 
To make TCAL comparable to the proposed method, in our experiments, we only use uncertainty and diversity criteria and ResNet18 for image classification. 
We use the same initial training set for TCAL as for ANNEAL and incorporate the same number of bits of information to the labeled training set in each AL iteration. Thus, we label $336 / \log_2{21} \approx 84$ images for TCAL, which corresponds to $5\%$ the training set. We train with Adam Optimizer and through cross-validation strategy set the learning rate as $10^{-3}$. 

\begin{figure}
    \centering
    \begin{tikzpicture}[scale = 0.78]
	\begin{axis}[
    height=7.2cm,
    width=11cm,
    line width=1pt,
    grid=both,
    grid style={line width=.1pt, draw=gray!10},
    major grid style={line width=.2pt,draw=gray!50},
    legend pos=south east,
    minor x tick num=4,
    minor y tick num=4,
    xlabel= {\normalsize Bits of Information},
    ylabel= {\normalsize mAP@5~(\%)},
    xmin=300,xmax=4150,
    ymin=40,ymax=97],
    
    \addplot+[name path=capacity] table [x=bits, y expr=(\thisrow{uncerainty + diversity} * 100), col sep=comma] {results_unbalanced.csv};\addlegendentry{ANNEAL};
    
    \addplot+[name path=capacity] table [x=bits, y expr=(\thisrow{uncerainty} * 100), col sep=comma] {results_unbalanced.csv};\addlegendentry{ANNEAL-U};
    
    \addplot+[name path=capacity] table [x=bits, y expr=(\thisrow{random} * 100), col sep=comma] {results_unbalanced.csv};\addlegendentry{Random};
    
    \addplot+[name path=capacity] table [x=bits, y expr=(\thisrowno{5} * 100), col sep=comma] {results_unbalanced.csv};\addlegendentry{ANNEAL (all samples)};
    \end{axis}
\end{tikzpicture}
    \caption{Analysis of the AL sampling strategies used within ANNEAL. }
    \label{Fig:results2}
\end{figure}
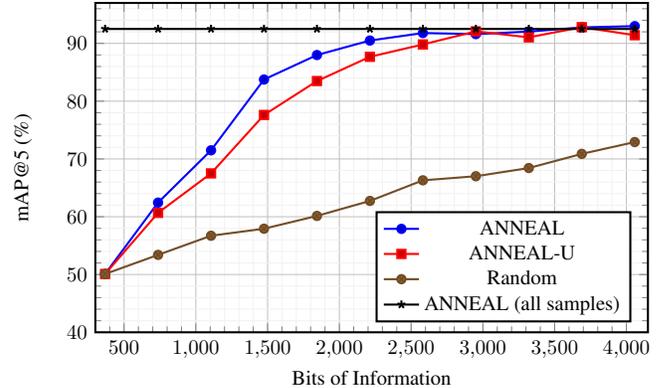

\begin{figure}
    \centering
    \vspace*{-3mm}
    \begin{tikzpicture}[scale = 0.78]
	\begin{axis}[
    height=7cm,
    width=11cm,
    line width=1pt,
    grid=both,
    grid style={line width=.1pt, draw=gray!10},
    major grid style={line width=.2pt,draw=gray!50},
    legend pos=south east,
    minor x tick num=4,
    minor y tick num=4,
    xlabel= {\normalsize Bits of Information},
    ylabel= {\normalsize mAP@5~(\%)},
    xmin=300,xmax=4100]
    
    \addplot+[name path=capacity] table [x=bits, y expr=(\thisrow{uncerainty + diversity} * 100), col sep=comma] {results_unbalanced.csv};\addlegendentry{ANNEAL};
    

    \addplot+[name path=capacity] table [x=bits, y expr=(\thisrow{uncerainty + diversity} * 100), col sep=comma] {results_images.csv};\addlegendentry{TCAL};

    \addplot+[name path=capacity] table [x=bits, y expr=(\thisrowno{5} * 100), col sep=comma] {results_unbalanced.csv};\addlegendentry{ANNEAL (all samples)};

    \addplot+[name path=capacity] table [x=bits, y expr=(\thisrowno{5} * 100), col sep=comma] {results_images.csv};\addlegendentry{TCAL (all samples)};

    \end{axis}

\end{tikzpicture}
    \caption{Comparison results between ANNEAL method and TCAL. }
    \label{LALA:results1}
\end{figure}
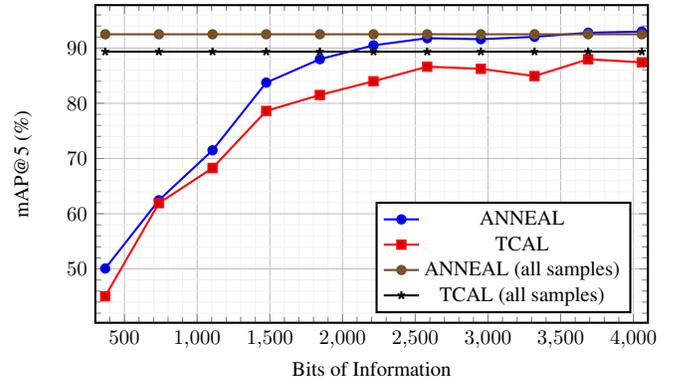

\begin{figure}[t]
    \centering
    \includegraphics[width=0.49\textwidth]{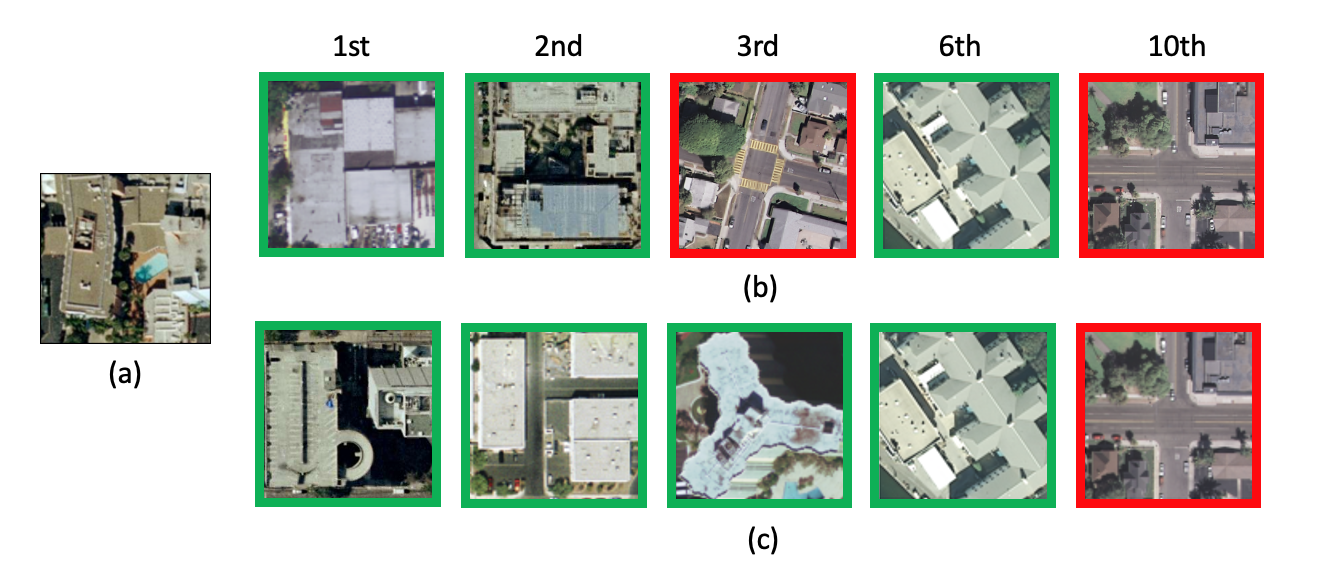}
    \caption{An image retrieval example: (a) query image annotated with the class label of \textit{building}; (b) images retrieved using the baseline TCAL; (c) images retrieved  using the ANNEAL. A green frame represents the same class label with the query image, while a red frame denotes a different class label with the query.}
    \label{Fig:retrieval_result}
\end{figure}

Fig.~\ref{LALA:results1} shows the retrieval performance  ANNEAL and TCAL in terms of mAP@5 versus the annotation cost in terms of bits of information. One can notice that our method outperforms TCAL at every number of bits of information utilized. The highest CBIR accuracy by using all pairs induced by all training images (ANNEAL all samples) is reached using only $2500$ bits of information. This corresponds to $34 \%$ of all LULC training images. The upper bound for LULC annotation is only reached with more than $4000$ bits, i.e. more than $54 \%$ of all training images. This clearly shows that  ANNEAL is more cost efficient in creating an optimized training set compared to TCAL.
Fig.~\ref{Fig:retrieval_result} shows a query image (see Fig. \ref{Fig:retrieval_result}a) and corresponding retrieved images of the TCAL method (see Fig.\ref{Fig:retrieval_result}b) and ANNEAL method (see Fig.\ref{Fig:retrieval_result}c) using $2500$ bits of information. From the figures, it can be observed that ANNEAL retrieves semantically more similar images to the query one compared to LULC annotation.
Overall, our results show that the proposed method is able to extract the most informative image pairs in $\mathcal{U}$ in a cost efficient manner while learning meaningful image features for CBIR. 

\section{Conclusion}
In this paper, we present a novel DML based AL method for CBIR that integrates the annotation of image pairs into the concept of one-bit supervision. By annotating image pairs as similar/dissimilar our approach significantly reduces the annotation cost compared to AL methods that are based on images labeled with LULC class labels. Experiments on the UCM dataset show that the proposed method is capable of creating a small and informative training set in an effective and cost-efficient way while achieving high CBIR performance. As a future development of this work, we plan to evaluate the informativeness of unlabeled image pairs directly on a feature space. This can lead to the more accurate selection of informative pairs and a decrease in the number of model parameters compared to using a binary classifier.


{\section{Acknowledgments}}
This work is supported by the European Research Council (ERC) through the ERC-2017-STG BigEarth Project under Grant 759764 and by the European Space Agency through the DA4DTE (Demonstrator precursor Digital Assistant interface for Digital Twin Earth) project. Julia Henkel's scholarship is funded by Google Research.

\bibliographystyle{IEEEbib}
{\small\bibliography{refs}}

\begin{thebibliography}{10}

\bibitem{Sumbul:2021}
G.~Sumbul, J.~Kang, and B.~Demir,
\newblock ``Deep learning for image search and retrieval in large remote
  sensing archives,''
\newblock in {\em Deep Learning for the Earth Sciences: A Comprehensive
  Approach to Remote Sensing, Climate Science and Geosciences}, chapter~11, pp.
  150--160. John Wiley \& Sons, Hoboken, NJ, USA, 2021.

\bibitem{surveyCBIR2021}
Y.~Li, J.~Ma, and Y.~Zhang,
\newblock ``Image retrieval from remote sensing big data: A survey,''
\newblock {\em Information Fusion}, vol. 67, pp. 94--115, 2021.

\bibitem{Feracatu_CBIRAL2007}
M.~Ferecatu and N.~Boujemaa,
\newblock ``Interactive remote-sensing image retrieval using active relevance
  feedback,''
\newblock {\em IEEE Transactions on Geoscience and Remote Sensing}, vol. 45,
  no. 4, pp. 818--826, 2007.

\bibitem{ALDemir2015}
B.~Demir and L.~Bruzzone,
\newblock ``A novel active learning method in relevance feedback for
  content-based remote sensing image retrieval,''
\newblock {\em IEEE Transactions on Geoscience and Remote Sensing}, vol. 53,
  no. 5, pp. 2323--2334, 2015.

\bibitem{schohn2000less}
G.~Schohn and D.~Cohn,
\newblock ``Less is more: Active learning with support vector machines,''
\newblock {\em Proc. Int. Conf. Mach. Learn}, vol. 2, no. 4, pp. 839--–846,
  2000.

\bibitem{hu2020onebitsupervision}
H.~Hu, L.~Xie, Z.~Du, R.~Hong, and Q.~Tian,
\newblock ``One-bit supervision for image classification,''
\newblock {\em Advances in Neural Information Processing Systems}, vol. 33, pp.
  501--511, 2020.

\bibitem{onebitactivequery2022}
Y.~Zhang, X.~Zhang, L.~Xie, J.~Li, R.~Qiu, H.~Hu, and Q.~Tian,
\newblock ``One-bit active query with contrastive pairs,''
\newblock {\em IEEE CCVPR}, pp. 9697--9705, 2022.

\bibitem{roy2018exploiting}
S.~Roy, S.~Paul, N.E. Young, and A.K Roy-Chowdhury,
\newblock ``Exploiting transitivity for learning person re-identification
  models on a budget,''
\newblock {\em Proceedings of the IEEE CVPR}, pp. 7064--7072, 2018.

\bibitem{contrastiveloss:2006}
R.~Hadsell, S.~Chopra, and Y.~LeCun,
\newblock ``Dimensionality reduction by learning an invariant mapping,''
\newblock {\em IEEE CCVPR}, vol. 2, pp. 1735--1742, 2006.

\bibitem{UCM_dataset}
Y.~Yang and S.~Newsam,
\newblock ``Geographic image retrieval using local invariant features,''
\newblock {\em IEEE Transactions on Geoscience and Remote Sensing}, vol. 51,
  no. 2, pp. 818--832, 2013.

\bibitem{branco2016survey}
P.~Branco, L.~Torgo, and R.~Ribeiro,
\newblock ``A survey of predictive modeling on imbalanced domains,''
\newblock {\em ACM computing surveys (CSUR)}, vol. 49, no. 2, pp. 1--50, 2016.

\end{thebibliography}
\vfill
\end{document}